\def\paperID{17009}
\def\confName{CVPR}
\def\confYear{2025}

\def\authorBlock{
    Souhail Hadgi$^{1,}$\thanks{\scriptsize Corresponding author: Souhail Hadgi (\texttt{hadgi@lix.polytechnique.fr})} \qquad
    Luca Moschella$^{2,}$\thanks{\scriptsize Currently at Apple.} \qquad
    Andrea Santilli$^2$ \\
    Diego Gomez$^1$ \qquad Qixing Huang$^4$ \qquad Emanuele Rodolà$^2$ \\
    Simone Melzi$^3$ \qquad Maks Ovsjanikov$^1$
    \\
    {\tt\small $^1$École polytechnique} \qquad \qquad \qquad
    {\tt\small $^2$Sapienza University of Rome} \\
    {\tt\small $^3$University of Milano-Bicocca} \qquad
    {\tt\small $^4$The University of Texas at Austin} \\
}

\newif\ifreview 
\newif\ifarxiv \newcommand{\arxiv}{\arxivtrue}
\newif\ifcamera 
\newif\ifrebuttal 

\arxiv 

\pdfoutput=1
\documentclass[10pt,twocolumn,letterpaper]{article}
\ifreview \usepackage[review]{cvpr} \fi
\ifarxiv \usepackage[pagenumbers]{cvpr} \fi
\ifrebuttal \usepackage[rebuttal]{cvpr} \fi
\ifcamera \usepackage{cvpr} \fi

\usepackage{graphicx}	
\usepackage{amsmath}	
\usepackage{amssymb}	
\usepackage{booktabs}
\usepackage{times}
\usepackage{microtype}
\usepackage{epsfig}
\usepackage{caption}
\usepackage{float}
\usepackage{placeins}
\usepackage{color, colortbl}
\usepackage{stfloats}
\usepackage{enumitem}
\usepackage{tabularx}
\usepackage{xstring}
\usepackage{multirow}
\usepackage{xspace}
\usepackage{url}
\usepackage{subcaption}
\usepackage{xcolor}
\usepackage[hang,flushmargin]{footmisc}

\ifcamera \usepackage[accsupp]{axessibility} \fi

% ...

\ifarxiv  \fi

\newcommand{\R}[1]{{%
    \textbf{%
        \ifstrequal{#1}{1}{\textcolor{red}{R#1}}{%
        \ifstrequal{#1}{2}{\textcolor{blue}{R#1}}{%
        \ifstrequal{#1}{3}{\textcolor{magenta}{R#1}}{%
        \ifstrequal{#1}{4}{\textcolor{teal}{R#1}}{%
                           \textcolor{cyan}{R#1}%
        }}}}%
    }%
}}

\usepackage{xr-hyper}

\makeatletter
\newcommand*{\addFileDependency}[1]{
  \typeout{(#1)}
  \@addtofilelist{#1}
  \IfFileExists{#1}{}{\typeout{No file #1.}}
}

\makeatother

\definecolor{cvprblue}{rgb}{0.21,0.49,0.74}
\usepackage[pagebackref,breaklinks,colorlinks,allcolors=cvprblue]{hyperref}
\usepackage[capitalize]{cleveref}
\crefname{section}{Sec.}{Secs.}
\crefname{table}{Table}{Tables}
\crefname{figure}{Fig.}{Figs.}

\ifarxiv \crefname{appendix}{App.}{Apps.}
\else \crefname{appendix}{Suppl.}{Suppls.} \fi

\usepackage{soul}
\usepackage{xcolor}

\begin{document}
\title{Escaping Plato's Cave: Towards the Alignment of 3D and Text Latent Spaces}
\author{\authorBlock}
\maketitle
\newcommand{\mypara}[1]{\vspace{1ex}\noindent\textbf{#1}~}

\begin{figure*}[ht!]
    \centering \includegraphics[width=\linewidth]{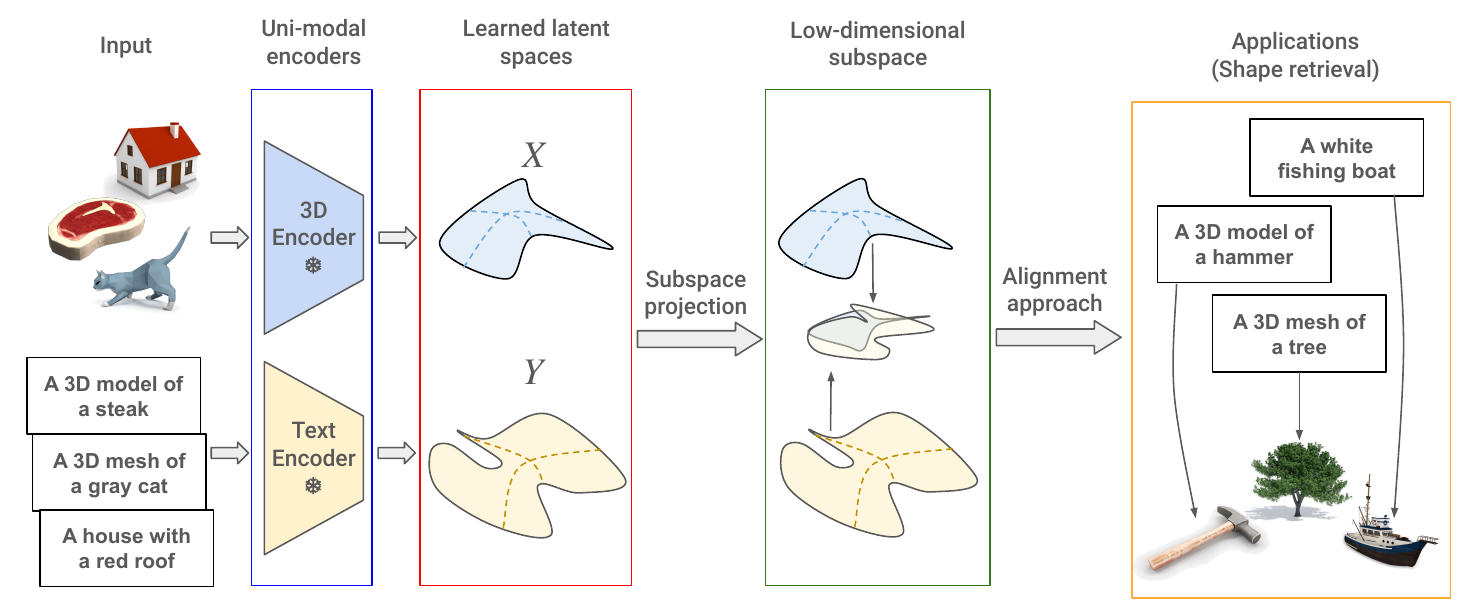}\caption{\label{fig:teaser} \textbf{A visual overview of the proposed approach.} is illustrated. From left to right: We begin with two distinct input collections—one consisting of 3D shapes and the other of textual prompts. In the blue box, independent, frozen uni-modal encoders map each modality into separate, high-dimensional latent spaces, shown in the red box. A dimensionality reduction procedure is applied to project these learned spaces into low-dimensional subspaces, represented in the green box. Finally, an alignment method registers the two low-dimensional subspaces, enabling cross-modal applications such as shape retrieval, with examples depicted in the yellow box.}
\end{figure*}

\begin{abstract}
Recent works have shown that, when trained at scale, uni-modal 2D vision and text encoders converge to learned features that share remarkable structural properties, despite arising from different representations. However, the role of 3D encoders with respect to other modalities remains unexplored. Furthermore, existing 3D foundation models that leverage large datasets are typically trained with explicit alignment objectives with respect to frozen encoders from other representations. In this work, we investigate the possibility of \emph{a posteriori} alignment of representations obtained from uni-modal 3D encoders compared to text-based feature spaces. We show that naive post-training feature alignment of uni-modal text and 3D encoders results in limited performance. We then focus on extracting subspaces of the corresponding feature spaces and discover that by projecting learned representations onto well-chosen lower-dimensional subspaces the quality of alignment becomes significantly higher, leading to improved accuracy on matching and retrieval tasks. Our analysis further sheds light on the nature of these shared subspaces, which roughly separate between semantic and geometric data representations. Overall, ours is the first work that helps to establish a baseline for post-training alignment of 3D uni-modal and text feature spaces, and helps to highlight both the shared and unique properties of 3D data compared to other representations. Our code and weights are available at \url{https://github.com/Souhail-01/3d-text-alignment}.
\end{abstract}
\section{Introduction}
\label{sec:intro}
Recent advances in multi-modal learning, particularly in vision-language models such as CLIP \cite{radford2021learning}, have sparked interest in extending these successes to the 3D domain. Most current approaches primarily focus on training 3D encoders through triplet-based learning with pre-trained 2D vision and language encoders \cite{liu2024openshape,xue2024ulip,zhou2023uni3d}, leveraging new large-scale datasets such as Objaverse \cite{deitke2022objaverse,objaverseXL}, and showing promising results in tasks such as zero-shot shape recognition. 

While CLIP \cite{radford2021learning} was trained with explicit alignment objectives between text and image representations, recent work has observed that \textit{even when trained independently}, vision and text encoders tend to exhibit significant similarities \cite{huh2024platonic,maniparambil2024vision}. In particular, the learned latent spaces of pure (uni-modal) text and vision encoders have similar proximity structure 
\cite{huh2024platonic} and can be aligned relatively easily \textit{after training} with a small number of known anchor correspondences 
\cite{maniparambil2024vision,moschella2023relative,cannistraci2023a}. Furthermore, the degree of similarity in learned features across modalities strongly correlates with the quality of performance in various downstream tasks. One prominent interpretation of these results is that given sufficient scale of training data and model complexity, different representations are converging to a shared underlying structure of the physical world. This has given rise to the Platonic Representation Hypothesis 
\cite{huh2024platonic}, where different representations are viewed as projections of reality on particular modalities.

At the same time, as the physical world is inherently (at least) 3-dimensional, a natural question is how the structure of uni-modal vision or text encoders relates to the features learned \textit{directly from} 3D data. This question raises several challenges: first, large-scale 3D datasets have only recently become available \cite{deitke2022objaverse,objaverseXL}; second, most existing 3D ``foundation models'' are trained with \textit{explicit alignment} objectives with respect to frozen 2D and text encoders, which limits the utility of post-training comparison. Finally, there is a lack of universally agreed-upon architectures and training objectives for 3D data, and many commonly-used architectures tend to have limited generalization power \cite{xie2020pointcontrast}.

In this work, we initiate the first study on the relation between 3D and language representations. We formulate the task of post-training alignment between a range of 3D and text encoders and study the accuracy and utility of several alignment strategies. 

Our first insight is that when trained on pure 3D data with self-supervised objectives, 3D encoders tend to lead to representations that align only very weakly to text-based representations. We believe that this observation already highlights the difficulty of uni-modal training and sheds light on the scarcity of ``pure'' 3D foundation models. 

Additionally, we reveal that while 3D and text latent spaces are not naturally aligned, effective cross-modal translation can be achieved through \textit{subspace projection and alignment}. Our key insight is that by identifying and operating in correlated subspaces, we can improve the latent space alignment of 3D and text encoders without the need for expensive joint training. To achieve this, we introduce a simple but effective approach that combines Canonical Correlation Analysis (CCA) and existing alignment approaches such as affine translation \cite{maiorca2024latent} and local CKA \cite{maniparambil2024vision}. This approach and our subsequent analysis extend previous works aimed at comparing learned feature spaces, but shows that careful \textit{subspace} alignment can reveal subtle but important similarities, which are otherwise obscured in global comparisons.
We provide a visual overview of the proposed approach in \cref{fig:teaser}.

To summarize, our main contributions are as follows:
\begin{enumerate}[topsep=1pt,itemsep=0ex,partopsep=1ex,parsep=1ex]
\item We extend the analysis of vision-text uni-modal latent space alignment to 3D uni-modal encoders and text, highlighting the limited similarity between these latent spaces and the low efficacy of current alignment approaches.
\item We propose an efficient approach for cross-modal alignment between text and 3D features that combines CCA for subspace selection and existing alignment methods. This approach improves alignment performance with minimal computational overhead, as demonstrated through experiments in matching and retrieval tasks. Our method establishes a baseline for 3D-text cross-modal understanding, providing an alternative to explicit multimodal pre-training for cross-modal tasks.
\item We observe a complementary structure between the spanned subspaces and the original feature spaces, enabling a distinction between geometric and semantic representations. 

\end{enumerate}

\section{Related Work}
\label{sec:related}

\mypara{Multi-modal representation learning.}
Multimodal representation learning has surged in recent years, driven by the success of image-language models \cite{radford2021learning, saharia2022photorealistic, li2022grounded, jia2021scaling} that enable seamless cross-modal applications. These models serve as the backbone for tasks spanning from text-based image retrieval to generating high-quality visuals from natural language prompts. By aligning visual and linguistic features in a shared latent space, these models have paved the way for advanced interactions between modalities, setting the stage for applications where visual and textual information are jointly processed and understood.

Building on these advancements, vision-language models have recently been adapted for 3D point cloud representation, where 3D-image-text triplets \cite{xue2023ulip, xue2024ulip, liu2024openshape, zhou2023uni3d} enable contrastive pre-training. These models leverage powerful techniques such as momentum contrast \cite{he2019momentum} to align representations across modalities, allowing point cloud encoders to be pre-trained in a multimodal context. Additionally, techniques that apply vision-language models directly to 2D projections of point clouds \cite{zhang2022pointclip, zhu2022pointclip} have expanded cross-modal applications to 3D, including zero-shot shape classification, where models trained on 2D-image-text data demonstrate strong performance on 3D-related tasks.

In the domain of multimodal synthesis, recent efforts in 2D and 3D model generation have leveraged diffusion models to tackle complex generation tasks. These models use robust priors, often trained on vast datasets, which allow them to create high-fidelity outputs in scenarios such as novel view synthesis and realistic 3D reconstructions from single RGB images \cite{liu2023zero1to3, xue2023human3diffusion}. By combining synthetic data with diffusion-based priors, these frameworks achieve impressive zero-shot generalization and geometry-consistent 3D synthesis, opening new avenues for applications where generating realistic 3D content from limited information is critical.

\mypara{Representational similarity.}
The study of representation similarity across neural networks has seen a significant rise in interest, spurred largely by seminal works originating in neuroscience and computational cognitive science. These fields have long been invested in understanding the nature and alignment of cognitive representations, providing a foundational basis that has influenced the current trajectory in machine learning \cite{sucho,moschella2024}. Based on these insights, various works \cite{mikolov2013exploiting, li2015convergent, morcos18, vulic2020all, bonheme2022variational, norelli2023asif, moschella2023relative} provide evidence of an intrinsic connection between independently trained networks; the similarity is especially notable among large-scale models, with works such as \cite{mikolov2013exploiting,9878514,mehta2022needgoodembeddingsextractor} exploring the phenomenon. 
In the computer vision and pattern recognition areas, the line of work on similarity-based representations, pioneered by Duin and Pekalska \cite{duin01}, has also been seminal. This line of research, which examines data through the lens of similarity rather than feature attributes, has provided robust frameworks for classification and clustering \cite{chen09}, enabling models to generalize across patterns and variations in complex datasets. 
Although mostly empirical in nature, these observations find theoretical support in the study of harmonics in neural networks weights \cite{Marchetti2023harmonics}, Independent Component Analysis \citep{roeder2021,hyvarinen2019,khemakhem2020,hyvarinen2016} and Independent Mechanism Analysis \citep{sliwa2022probing,gresele2021independent,ghosh2023independent}. These works suggest that, when capturing the same underlying data generative factors, deep learning models may converge towards similar structures despite their complexity and non-linearity. 

\mypara{Latent space alignment.}
More recently, a range of approaches has been developed to {\em align} latent spaces within the same modality \cite{crisostomi2023from,cannistraci2024from,fumero2024latent}, as well as across different modalities \cite{9150585,9878889,maiorca2024latent}, particularly between visual and textual domains. Techniques such as Procrustes analysis \cite{wang2008manifold} and several similarity metrics \cite{klabunde2024similarityneuralnetworkmodels, SHAHBAZI2021118271, tang2020similarityneuralnetworksgradients, williams2021generalized}, including centered kernel alignment (CKA) \cite{kornblith2019similarity, davari2022reliability, maniparambil2024vision}, have proven instrumental in aligning representations. These methods offer various strategies for quantifying similarity between feature spaces, allowing us to examine cross-modal interactions at a deeper level. HHowever, these methods often focus on aligning entire latent spaces, potentially missing meaningful similarities confined to specific subspaces.

Our approach builds upon CCA \cite{hotelling1992relations, raghu2017svcca}, a pivotal tool in pattern recognition and multi-view learning applications \cite{guo2019canonical, rupnik2010multi}. This technique identifies maximally correlated subspaces, enabling more refined alignment across modalities by isolating core, mutually relevant components. Leveraging CCA, alignment methods can extend into complex domains, such as connecting 3D and textual latent spaces, as we demonstrate in the following.

\section{Method}
\label{sec:method}
We compare the similarity between latent (feature) spaces of various 3D and text encoders, introducing a new approach that builds on existing alignment methods to improve their effectiveness. Below, we outline the tools used to measure and align these latent spaces.
\subsection{Preliminaries}\label{sec:bg}
\mypara{Centered Kernel Alignment.}
CKA is a similarity measure frequently used in recent studies \cite{kornblith2019similarity, davari2022reliability} to compare representations in neural network feature spaces. Given feature matrices $\mathbf{X} \in \mathbb{R}^{n \times p}$ and $\mathbf{Y} \in \mathbb{R}^{n \times q}$, we apply kernel functions $k$ and $l$ to obtain kernel representations $\mathbf{K} = k(\mathbf{X}, \mathbf{X}) \in \mathbb{R}^{n\times n}$ and $\mathbf{L} = l(\mathbf{Y}, \mathbf{Y})\in \mathbb{R}^{n\times n}$. CKA is defined as:
\begin{equation}
    \text{CKA}(\mathbf{K}, \mathbf{L}) = \frac{\text{HSIC}(\mathbf{K}, \mathbf{L})}{\sqrt{\text{HSIC}(\mathbf{K}, \mathbf{K})\text{HSIC}(\mathbf{L}, \mathbf{L})}}
\end{equation}
where HSIC is the Hilbert-Schmidt Independence Criterion \cite{gretton2005measuring} and can be written as 
\begin{equation}
\text{HSIC}(\mathbf{K}, \mathbf{L}) = \frac{1}{(n-1)^2}\text{tr}(\mathbf{K}\mathbf{H}\mathbf{L}\mathbf{H}) \,,
\end{equation}
where $\mathbf{H} = \mathbf{I} - \frac{1}{n}\mathbf{1}\mathbf{1}^\top $.

\mypara{Canonical Correlation Analysis.}
CCA \cite{hotelling1992relations,raghu2017svcca} is a statistical method that finds linear projections of two datasets, maximizing their correlation in a shared latent space.

Formally, given two sets of zero-centered variables \( \mathbf{X} \in \mathbb{R}^{n \times d_1} \) and \( \mathbf{Y} \in \mathbb{R}^{n \times d_2} \), CCA seeks two projection matrices \( \mathbf{W}_{\mathbf{X}} \in \mathbb{R}^{d_1 \times k} \) and \( \mathbf{W}_{\mathbf{Y}} \in \mathbb{R}^{d_2 \times k} \) that map \( \mathbf{X} \) and \( \mathbf{Y} \) into a common \( k \)-dimensional space, maximizing the correlation between the projections \( \mathbf{X} \mathbf{W}_{\mathbf{X}} \) and \( \mathbf{Y} \mathbf{W}_{\mathbf{Y}} \). The optimization can be expressed as:
\begin{equation}
\max_{\mathbf{W}_{\mathbf{X}}, \mathbf{W}_{\mathbf{Y}}} \operatorname{corr}(\mathbf{X} \mathbf{W}_{\mathbf{X}}, \mathbf{Y} \mathbf{W}_{\mathbf{Y}}) \,,
\end{equation}
where \( \operatorname{corr}(\cdot, \cdot) \) represents the correlation between the projected variables.

\subsection{Alignment approaches}\label{sec:bg}
Recent developments in latent space alignment have introduced various methods. In this work, we examine both the affine transformation approach in \cite{maiorca2024latent} and the CKA-based matching approach from \cite{maniparambil2024vision}, observing their limited effectiveness in 3D-text latent space alignment, and propose a method to address these limitations.

\mypara{Latent Space Translation through Affine Transformation~\cite{maiorca2024latent}.} 
It is possible to estimate an affine transformation $T$ that maps one latent space $\mathcal{X}$ onto another latent space $\mathcal{Y}$ such that $T(\mathbf{x}) = \mathbf{Rx} + \mathbf{b},\forall \mathbf{x}\in \mathcal{X}$. To compute $T$, we assume access to an \textit{anchor} subset consisting of ground-truth paired samples from both latent spaces. These anchors enable us to determine $T$ by optimizing through gradient descent or, if $\mathbf{b} = {\mathbf{0}}$ by minimizing the least squares error. It assumes that $\mathcal{X}$ and $\mathcal{Y}$ share the same dimensionality and are normalized to zero mean and unit variance. These constraints can be enforced by zero-padding the smaller latent space.

\mypara{Local CKA-based retrieval and matching~\cite{maniparambil2024vision}.} 
The CKA metric computed between two sets of features is sensitive to ordering and maximized when ground-truth pairs are aligned, this insight can be used to match unseen data by including them in a well-aligned anchors set. Formally, starting from aligned set of features $\mathbf{X}_A$ and $\mathbf{Y}_A$, we compute for a query pair $(\mathbf{x}_q,\mathbf{y}_q)$ its local CKA defined as: 
\begin{equation}
\text{localCKA}(\mathbf{x}_q, \mathbf{y}_q) = \text{CKA}(\mathbf{K}_{[\mathbf{X}_A,\mathbf{x}_q]}, \mathbf{K}_{[\mathbf{Y}_A,\mathbf{y}_q]}) \,,
\end{equation}
where $[\mathbf{X}, \mathbf{x}]$ denotes the column-wise concatenation of matrix $\mathbf{X}$ and vector $\mathbf{x}$. Local CKA calculates similarity in a pairwise manner, accounting for anchor alignment while being sensitive to ordering among query pairs. We introduce all possible query pairs, where correctly matched pairs exhibit the highest local CKA.

\subsection{Proposed method.}\label{sec:ours}

Our goal is to reliably align the latent spaces of pre-trained 3D and text encoders. Given a dataset of $n$ caption-point cloud pairs, each embedded in the corresponding feature space, and represented as matrices $\mathbf{X} \in \mathbb{R}^{n \times p}$ and $\mathbf{Y} \in \mathbb{R}^{n \times q}$ respectively, we first select a subset of anchor pairs that will guide our translation process, denoted as $\mathbf{X}_A  \in \mathbb{R}^{n_{A} \times p}$ and $\mathbf{Y}_A \in \mathbb{R}^{n_{A} \times q}$. Building upon the mathematical foundations introduced in Section \ref{sec:bg}, we develop a pipeline that combines dimensionality reduction with the previously introduced alignment methods.

\mypara{Common Subspace Projection.}
In our work, we show that 3D and text latent spaces can effectively be aligned in low-dimensional connected subspaces. We begin by applying CCA to the anchor pairs to identify a shared $k$-dimensional subspace (with $k<p,q$) that connects point cloud and text latent spaces. This yields projection matrices $\mathbf{W}_{\mathbf{X}^A} \in \mathbb{R}^{p \times k}$ and $\mathbf{W}_{\mathbf{Y}^A} \in \mathbb{R}^{q \times k}$. All samples are then projected into this reduced space:
\begin{equation}
\mathbf{X}^r = \mathbf{X}\mathbf{W}_{\mathbf{X}_A}, \quad \mathbf{Y}^r = \mathbf{Y}\mathbf{W}_{\mathbf{Y}_A} \,.
\end{equation}
In particular, we refer to the reduced anchors as ${\mathbf{X}_A^r \in \mathbb{R}^{n_{A} \times k} \text{ and } \mathbf{Y}_A^r}\in \mathbb{R}^{n_{A} \times k}$.

Our experiments show that projecting 3D and text latent spaces into a lower-dimensional, shared subspace improves alignment by isolating features that are highly correlated across modalities

\mypara{Alignment of projected latent spaces.}
Given the projected latent spaces, we can apply either of the previously described alignment methods in the reduced space. For the affine transformation approach, we learn a mapping between $\mathbf{X}^r$ and $\mathbf{Y}^r$ using the projected anchor pairs ${\mathbf{X}_A^r, \mathbf{Y}_A^r}$ to optimize the transformation parameters $\mathbf{R}$ and $\mathbf{b}$:
\begin{equation}
T(\mathbf{X}^r) = \mathbf{RX}^r + \mathbf{b},
\end{equation}
Alternatively, we can employ the local CKA-based matching approach in the projected space. For a query pair $(\mathbf{x}_q^r, \mathbf{y}_q^r)$, we compute its local CKA using the projected anchor sets:

\begin{equation}
\text{localCKA}(\mathbf{x}_q^r, \mathbf{y}_q^r) = \text{CKA}(\mathbf{K}_{[\mathbf{X}_A^r,\mathbf{x}_q^r]}, \mathbf{K}_{[\mathbf{Y}_A^r,\mathbf{y}_q^r]}) ,
\end{equation}

\begin{table*}[t]
\centering
\resizebox{\linewidth}{!}{%
\begin{tabular}{c|c|cc|cc|cc}
\hline
\textbf{Method} & \textbf{3D Encoder} & \multicolumn{2}{c|}{\textbf{CLIP}} & \multicolumn{2}{c|}{\textbf{RoBERTa}} & \multicolumn{2}{c}{\textbf{BERT}} \\ 
                &                     & Matching accuracy & Top-5 retrieval & Matching accuracy & Top-5 retrieval & Matching accuracy & Top-5 retrieval \\ \hline
\multicolumn{8}{c}{\textbf{Multi-modal 3D encoder}} \\ \hline
Affine + Subspace Projection     & OpenShape           &  67.6            & 85.6            & 55.2            & 75.8            & 45.0            & 70.6            \\ 
Affine + Subspace Projection     & ULIP-2              & 65.6            & 85.2            & 47.2            & 70.6            & 35.2            & 59.8            \\
Affine + Subspace Projection     & Uni3D              & 61.4            & 81.6            & 45.8            & 63.8            & 34.4            & 47.2            \\ \hline
\multicolumn{8}{c}{\textbf{Uni-modal 3D encoder}} \\ \hline
Affine       & PointBert           & 15.8            & 23.6            & 7.8            & 15.6            & 6.4             & 13.4            \\ 
Affine       & SparseConv              & 11.0            & 34.4            & 6.0             & 20.0            & 4.2             & 16.2            \\ 
Affine       & Pointnet            & 18.4            & 21.8            & 8.0             & 10.2            & 9.6             & 12.2            \\ \hline
Affine + Subspace Projection (Ours)       & PointBert           & \textbf{30.8}            & 42.2            & \textbf{23.2}            & 28.4            & 15.6             & 18.2            \\ 
Affine + Subspace Projection (Ours)       & SparseConv              & 21.4            & 45.6            & 19.2             & 16.8            & 15.8             & 15.0            \\ 
Affine + Subspace Projection (Ours)       & Pointnet            & 25.2            & 36.6            & 22.4             & 20.8            & \textbf{16.6}             & 16.0  \\ \hline
Local CKA       & PointBert           & 5.8            & 15.2            & 1.8            & 1.4            & 1.0             & 4.39            \\ 
Local CKA       & SparseConv              & 3.4            & 13.6            & 1.79             & 1.6            & 0.6            & 3.8            \\ 
Local CKA       & Pointnet            & 6.6            & 18.0            & 2.4             & 2.0            & 1.0            & 5.0  \\ \hline
Local CKA + Subspace Projection (Ours)       & PointBert           & 29.4            & \textbf{60.19}            & 17.0            & \textbf{42.4}            & 15.0             & \textbf{37.2}            \\ 
Local CKA + Subspace Projection (Ours)       & SparseConv              & 19.0            & 56.9            & 15.8             & 34.0            & 10.0             & 30.8           \\ 
Local CKA + Subspace Projection (Ours)       & Pointnet            & 26.8            & 53.0            & 18.6             & 40.2            & 14.3             & 38 \\ \hline
\end{tabular}%
}
\caption{\textbf{Matching and retrieval performance across 3D and text encoders using different alignment approaches.} We use 30,000 anchors for subspace projection and affine transformation approaches, and 1,000 anchors for local CKA. A query set of 500 is uniformly sampled, with results averaged over 3 different seeds. The subspace dimension is fixed at 50. Our approach (Ours) consistently demonstrates improved matching and retrieval performance, with multi-modal 3D encoders setting the upper bound for performance. Additional top-$k$ retrieval metrics are provided in the supplementary.}
\vspace{-0.05in}
\label{tab:performance}
\end{table*}

\section{Experimental setup}
\label{sec:exps}

\subsection{Pre-training Dataset}
\label{subsec:pretraining_dataset}
Prior works in point cloud pre-training, particularly within uni-modal frameworks, have relied heavily on ShapeNet \cite{chang2015shapenet}, a dataset of 51,300 annotated 3D synthetic shapes spanning 55 categories. ShapeNet has been instrumental in advancing foundational methods, yet it remains limited by the relatively narrow scope of categories. With the release of Objaverse \cite{deitke2022objaverse}, which includes over 800,000 shapes across diverse real-world categories, a new standard for large-scale representation learning in the 3D domain has emerged. Objaverse's extensive shape diversity makes it ideal for both uni-modal and multi-modal learning. Despite these advantages, there is a lack of works that explore uni-modal pre-training specifically on Objaverse.

\subsection{Encoders}
\label{subsec:encoders}
We explore both multi-modal and uni-modal 3D and text encoders across varying levels of model complexity.

\mypara{Multi-modal 3D Encoders.} For multi-modal pre-training, we use {OpenShape, ULIP-2 and Uni3D} \cite{liu2024openshape} pre-trained models, trained on point cloud, image, and text triplets of Objaverse with a contrastive pre-text task to align 3D encoders with frozen CLIP encoders. We adopt the Point-BERT-based variant of each model \cite{yu2021pre}. We primarily focus on OpenShape for simplicity but generalize results to ULIP-2 and Uni3D.

\mypara{Uni-modal 3D Encoders.} For the uni-modal setup, we pre-train PointBERT on Objaverse using its original pretext tasks, which include masked point reconstruction and a uni-modal contrastive loss to encourage robust shape representation. We also explore two additional architectures: a Sparse convolution (MinkowskiNet \cite{choy20194d}) model and a simpler architecture in PointNet \cite{qi2017pointnet}, each pre-trained using a shape-level contrastive learning method \cite{hadgi2024supervise}. This approach contrasts different partial views of input shapes. Across all 3D encoders in this setup, we fix the latent dimension at 512 to maintain consistency in representation space comparisons.

\mypara{Text encoders.} We use the text encoder from OpenCLIP ViT-bigG-14 \cite{ilharco_gabriel_2021_5143773}, chosen to match text encoder used in OpenShape. Additionally, we examine alignment with purely uni-modal text encoders by including BERT \cite{devlin2018bert} and RoBERTa \cite{liu2019roberta}, we also evaluate the alignement with T5 \cite{raffel2020exploring} in the supplementary. 

Across all pre-training setups, parameters are kept consistent to facilitate direct comparisons. We detail the technical details in the supplementary.

\subsection{Downstream tasks}
\label{subsec:downstream_tasks}

We evaluate our alignment approach on Objaverse-LVIS \cite{gupta2019lvis}, a human-verified test subset of Objaverse that contains 1,156 object categories. This subset is specifically reserved for evaluation and is unseen during pre-training. The captions are generated with Cap3D \cite{luo2024scalable}, which provides enhanced descriptive text for each 3D shape.
Our evaluation framework consists of two main tasks: matching and retrieval. The matching task involves finding the correct permutation of captions for perfect matching given a shuffled set of query images and their corresponding captions; we utilize the linear sum assignment approach to perform this task \cite{kuhn1955hungarian}. For the retrieval task, the model must identify the correct 3D object from the query set based on a text caption. These tasks are particularly effective in measuring the cross-modal capabilities of encoders, and have been evaluated in prior Vision-Text studies \cite{maniparambil2024vision}. While our main results emphasize the matching and top-5 retrieval tasks, we provide evaluation of top-1 and top-10 retrieval metrics in the supplementary.

\begin{figure}[tp]
    \centering
    \includegraphics[width=\linewidth]{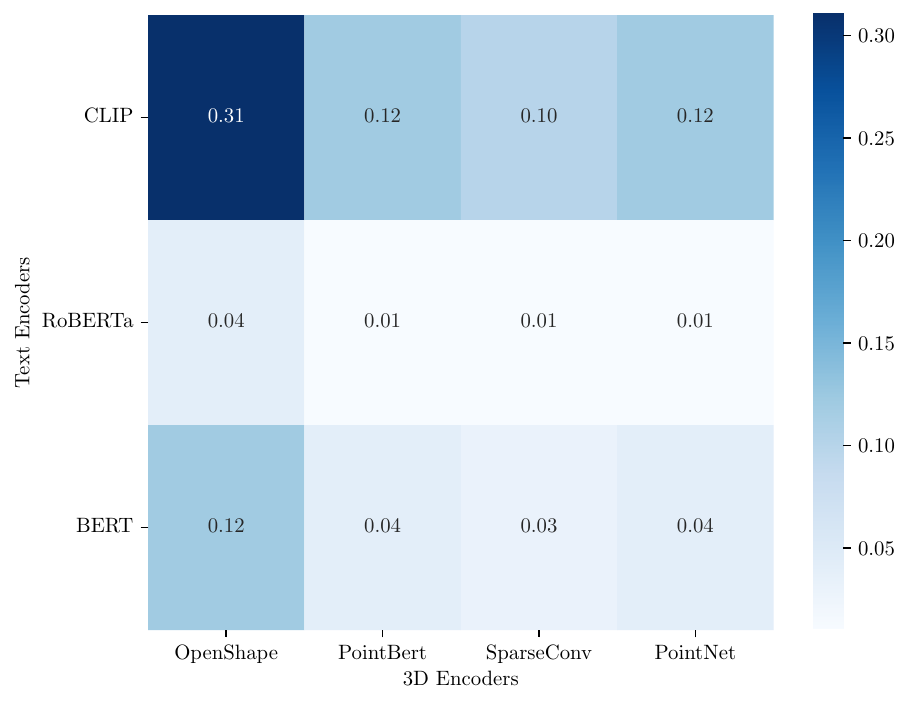}
    \caption{\textbf{Linear CKA scores between text and 3D encoders without alignment.} Higher scores reflect stronger alignment between encoder pairs, with the strongest alignment observed between multi-modal 3D encoders (OpenShape) and CLIP text encoder due to their shared training on aligned representations. Uni-modal 3D encoders show significantly lower alignment with text encoders, although slightly higher with the CLIP text encoder.}
    \label{fig:cka_pre}
\end{figure}

\begin{figure}[tp]
    \centering
    \includegraphics[width=\linewidth]{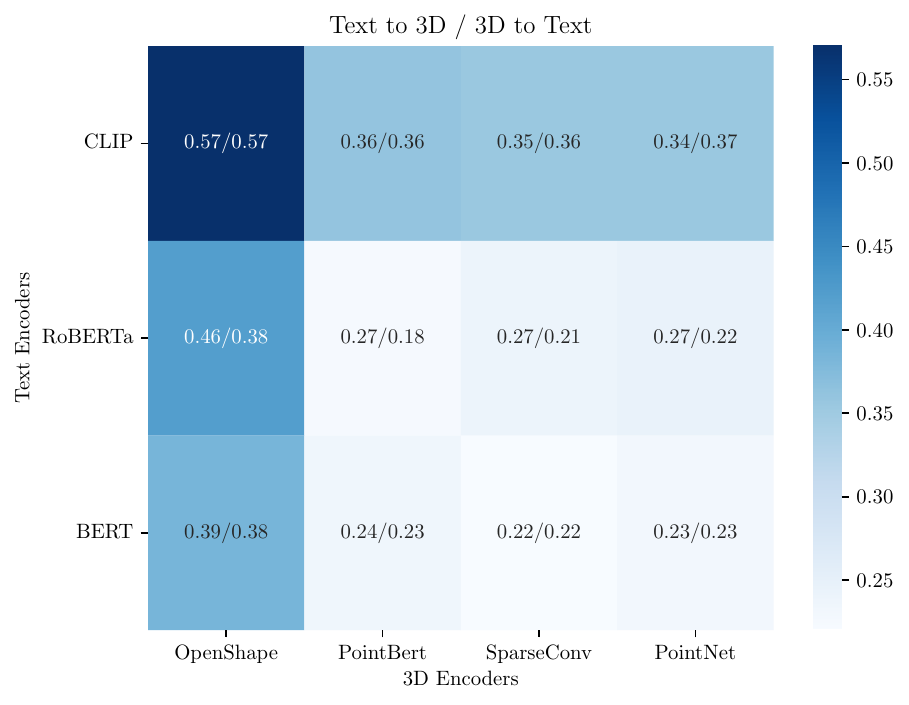}
    \caption{\textbf{Linear CKA scores between text and 3D encoders after affine translation.} Affine translation results in a consistent increase in similarity for both 3D-to-text and text-to-3D directions.}
    \label{fig:cka_post}
\end{figure}


\section{Results}
\label{sec:results}

\subsection{Are 3D and Text Latent Spaces similar ?}

To evaluate the inherent similarity between 3D and text feature spaces without alignment, we compute linear CKA scores for both unimodal and multimodal encoders. The results are shown in~\cref{fig:cka_pre}.

\mypara{Comparison of 3D-Text and Vision-Text alignment.} Prior work \cite{huh2024platonic} reports CKA alignment values ranging from approximately 30\% to 48\% between different uni-modal vision and text encoders (see Figure 13 in \cite{huh2024platonic}). In stark contrast, we find that the default alignment between 3D and text latent spaces is significantly weaker, with a maximum score of 0.12 observed for the uni-modal PointBERT and CLIP pair. This substantial gap underscores a key insight: unlike vision and text encoders, 3D encoders did not converge to structures similar to text.

\mypara{Alignment favors multi-modality.}  We observe that 3D multi-modal encoders demonstrate the highest CKA score with text encoders, particularly with CLIP's text encoder. This behavior is expected, given that 3D multi-modal encoders are explicitly trained to align with CLIP latent spaces, which share the same textual modality. Even among uni-modal 3D encoders, alignment with the CLIP text encoder is notably higher (e.g. 0.12 for PointBERT-CLIP) compared to alignment with uni-modal text encoders like RoBERTa (e.g., 0.04 for PointBERT-RoBERTa). This suggests that the visual understanding embedded in CLIP’s text encoder extends beyond image and text domains to include 3D representations, despite the lack of explicit alignment during pre-training.

\subsection{Latent Space Alignment results}

We evaluate the performance of the alignment approaches outlined in \cref{sec:method} using matching and retrieval tasks, and analyze their effectiveness in aligning 3D and text encoders. Unless otherwise specified, we fix the subspace dimension to $d = 50$ and the number of anchors to $30,000$. For downstream tasks, we uniformly sample a query set of size 500 and average results over 3 different seeds. For the affine transformation approach, we present results for the text-to-3D direction, noting that similar performance is achieved in the 3D-to-text direction as seen in \cref{fig:cka_post}.

\mypara{Existing approaches enable limited alignment} We first assess whether previous successful approaches—affine transformation and local CKA—can achieve meaningful 3D-text alignment. As shown in \cref{tab:performance}, both methods yield modest alignment improvements over the unaligned baseline, where uni-modal 3D encoders start near zero in matching and retrieval tasks. These results suggest a small alignment shift. However, even with this improvement, alignment remains significantly lower than the alignment achieved in vision-text benchmarks \cite{maniparambil2024vision}. This finding hints at the limits of uni-modal 3D encoders in achieving similar Vision-Text alignment performance, and might necessitate a different approach to align their latent space with text encoders.

\begin{figure}[tp]
  \centering
  \begin{subfigure}{0.236\textwidth}
    \includegraphics[width=1\linewidth]{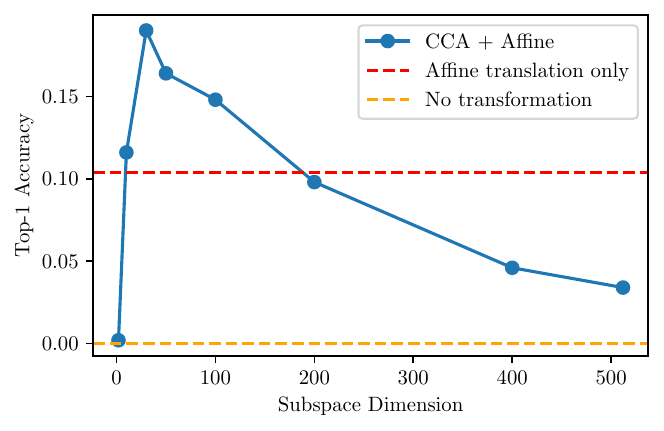}
  \end{subfigure}
  \begin{subfigure}{0.236\textwidth}
    \includegraphics[width=1\linewidth]{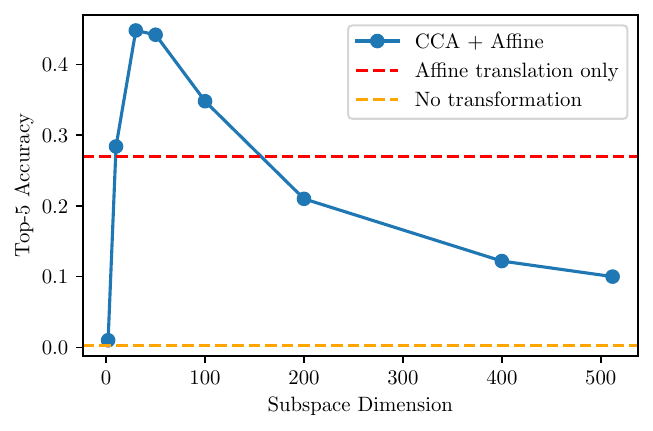}
  \end{subfigure}
  \caption{\textbf{Impact of subspace dimensionality on retrieval performance.} Comparison of three approaches: our proposed CCA + affine translation method (blue), affine translation without subspace projection (red), and baseline feature space alignment without transformation (orange). Results are shown for the uni-modal PointBERT and CLIP text encoder, with generalizations to other encoders provided in the supplementary.}
  \label{fig:dimension_ablation}
\end{figure}

\mypara{Importance of subspace projection (Ours).} While aligning the latent spaces of 3D and text encoders achieved some success with existing methods, results remained well below vision-text alignment benchmarks. Motivated by this limitation, we propose aligning lower-dimensional subspaces, based on the hypothesis that 3D and text representations might intersect within a shared latent subspace. Using a validation set, we report in \cref{fig:dimension_ablation} the impact of subspace dimension, obtained through our CCA approach (see \cref{sec:ours}) combined with the affine transformation on the top-5 retrieval accuracy of uni-modal PointBert. While affine alignment alone yields better performance at higher dimensions, our subspace projection approach significantly outperforms it when the subspace dimension is reduced. This finding indicates that alignment quality is optimized within carefully chosen lower-dimensional spaces, reinforcing our hypothesis that 3D-text alignment is more successful within targeted subspaces. The results in \cref{tab:performance} further validate this approach, showing a substantial increase in matching and retrieval performance across all 3D and text encoder pairs, outperforming both alignment approaches when they operate on the original latent space.

\mypara{Multi-modal encoders as an upper bound of performance.} We include 3D multi-modal encoders for two main reasons: (1) As an upper bound for alignment performance, (2) To study how alignment degrades across different text encoders, including those that were not used during pre-training. In this context, the inclusion of 'Affine + Subspace Projection' multi-modal results in \cref{tab:performance} shows how using our approach significantly narrows the gap between multi-modal and uni-modal performance. However, to illustrate point 1. above, we also measured with cosine similarity the alignment of OpenShape (a multi-modal 3D encoder) and the same CLIP text encoder used during pre-training, achieving 0.94 for top-5 retrieval accuracy, an expected result since 3D and text encoders are explictly aligned during pre-training.

\mypara{3D encoder complexity's low impact on alignment.} PointBERT outperforms the more complex SparseConv among uni-modal 3D encoders, suggesting that increasing model complexity alone does not guarantee improved alignment in 3D-text tasks. Surprisingly, even PointNet—a relatively simple model—achieves similar alignment scores, showing that factors other than model complexity may play a pivotal role in 3D-text alignment. This observation contrasts with vision-text alignment results \cite{huh2024platonic, maniparambil2024vision}, where complex models typically leverage large datasets more effectively. Our findings thus indicate a distinctive aspect of 3D-text alignment: model simplicity does not impede, and may even aid, the interpretability and compatibility of learned features for cross-modal alignment.

\mypara{Different alignment techniques have different strengths.} Our experiments reveal that different alignment techniques offer complementary strengths across tasks. For instance, affine transformation proves particularly effective for matching tasks across all 3D encoders, while local CKA shows superior performance in top-5 retrieval accuracy. This suggests that while some methods excel in precision tasks, others might better capture broader semantic nuances, making them more suitable for retrieval. Together, these observations imply that a hybrid approach could leverage the unique strengths of each method, opening up promising directions for future cross-modal applications.

\begin{figure}[tp]
    \centering
    \includegraphics[width=\linewidth]{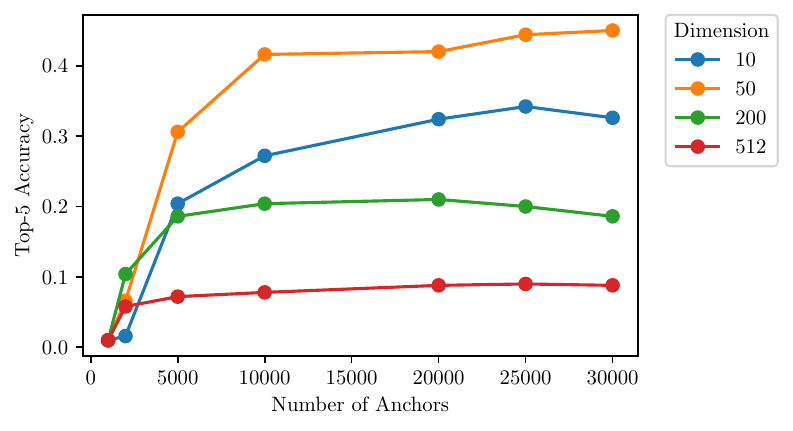}
    \caption{\textbf{Effect of anchor set size on retrieval performance.} Validation set results with different subspace dimensions show that retrieval performance improves as the anchor subset size increases but eventually reaches a plateau. Results are shown for the uni-modal PointBERT and CLIP text encoder.}
    \label{fig:anchors_ablation}
\end{figure}

\mypara{Scaling of our approach.} In \cref{fig:anchors_ablation}, we explore the scalability of our approach by analyzing how performance responds to increasing the number of anchors. Notably, our subspace projection method scales effectively with anchor count, reaching a plateau before requiring the full dataset (over 800,000 shapes). This scalability highlights the approach’s efficiency in learning robust 3D-text mappings with a limited subset of anchor pairs. However, we observe a leveling off in performance gains beyond a certain anchor count, likely due to the constraints imposed by the low-dimensional subspace. This suggests that while our approach is computationally efficient and data-efficient, its reliance on a fixed lower-dimensional subspace could limit its adaptability to larger, more diverse datasets.

\begin{figure}[tp]
    \centering
    \includegraphics[width=\linewidth]{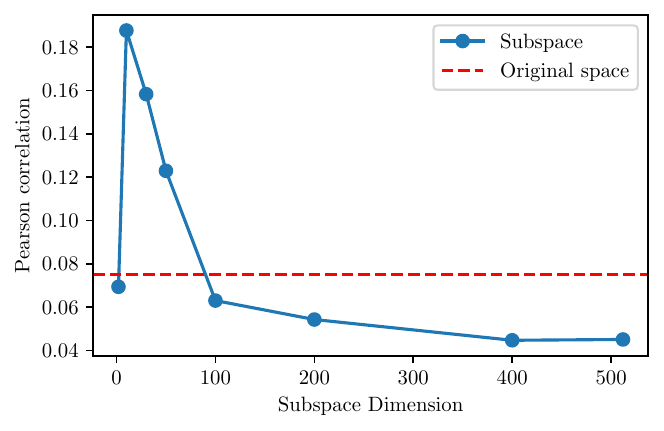}
    \caption{\textbf{Pearson correlation between shape query chamfer distances and pairwise distances in the projected text latent subspace.} We observe a higher Pearson correlation in the projected text latent subspace with optimal subspace dimension. Results are shown for the uni-modal PointBERT and CLIP text encoders.}
    \label{fig:pearson}
\end{figure}

\begin{figure}[tp]
    \centering
    \includegraphics[width=\linewidth]{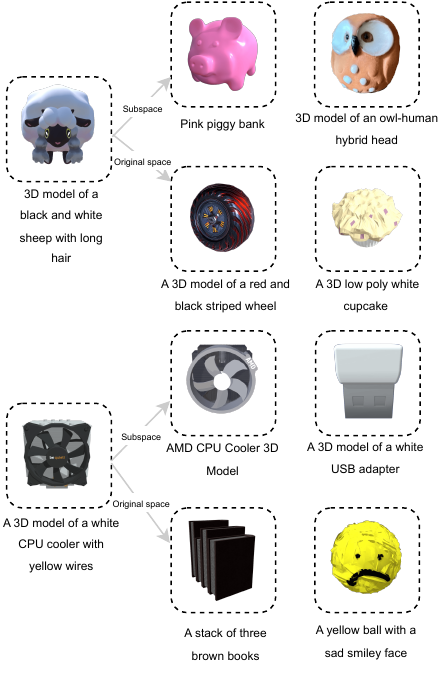}
    \caption{\textbf{Shape retrieval comparison between original and reduced latent spaces.} For a given query shape, we retrieve the closest match based on cosine similarity. Results demonstrate a higher semantic understanding in the reduced 3D latent subspace compared to the original latent space. Shown for uni-modal PointBERT and CLIP.}
    \label{fig:qualitative}
\end{figure}

\subsection{Geometries vs. semantics.}

Our quantitative evaluation shows that low-dimensional subspace projection significantly improves latent space alignment in the 3D-Text setting. To better understand the characteristics of these subspaces spanned relative to the original spaces, we analyze the increase in semantic and geometric knowledge within the projected spaces.

\mypara{Increased geometric awareness of the text latent subspace.}
To quantify the increase in geometric awareness within the projected text subspace, we compute the Pearson correlation between the Chamfer distances of a query set of 500 shapes and the pairwise distances between feature vectors within the text subspace and compare these results to those obtained from the original text latent space. Results in \cref{fig:pearson} show that the optimal subspace dimension yields a high correlation with Chamfer distances, signaling an increase in the sensitivity of the text encoder to geometric properties when projected into this specific subspace.

\mypara{Increased semantic understanding of the 3D latent subspace.}
To assess the improved semantic capacity of the 3D subspace, we conducted a qualitative analysis of shape retrieval performance in both the reduced and the original latent spaces (\cref{fig:qualitative}). Given a set of query shapes, we observe that the reduced subspace retrieves shapes with stronger semantic similarity (e.g., retrieving animals for a query sheep shape), while the original latent space primarily favors geometric resemblance (e.g retrieving stacked square shapes when querying a \textit{'3D model of a CPU'}). This shift suggests that our subspace projection method enables the 3D encoder to capture semantic relationships that are absent in the full latent space. This effectively enables cross-modal applications and explains its higher matching and retrieval performance.

\section{Conclusion, Limitations and Future Work}
\label{sec:conclusion}

In this work, we present the first study investigating latent space alignment between 3D and text pre-trained encoders. Building on the hypothesis that these modalities share semantic connections within lower-dimensional subspaces, we propose an effective approach combining CCA projection with affine transformation estimation to translate between modalities' latent spaces.
Our empirical results show that optimal cross-modal performance is achieved through low-dimensional subspace projection, and our method successfully improves alignment across diverse 3D and text uni-modal encoders. While CLIP-based multi-modal encoders establish performance upper bounds, we enable significant cross-modal capabilities in uni-modal encoders previously limited to single-modality tasks. We also demonstrate that semantic understanding can be extracted from geometry-aware latent spaces of uni-modal 3D encoders. 

Although our work focused on Objaverse, which is the first large-scale 3D dataset, it would be interesting to consider how scaling on Objaverse-XL \cite{objaverseXL} would affect the alignment quality between 3D and text encoders. Moreover, in this work we do not distinguish object-level vs. scene-level annotations, and decomposing objects or scenes into their composing blocks could shed light onto the compositionality of the learned representations. Finally, the subspace alignment method that we introduce in this work can be broadly applicable to other representations as well. In the future, we plan to use it to investigate the limitations of alignment observed in other representations. In particular, even when trained at significantly higher data scales, images and text representations do not \textit{align perfectly} \cite{huh2024platonic}, and it would be interesting to reveal the unique and complementary nature of different modalities via subspace analysis.

\mypara{Acknowledgements} Parts of this work were supported by the ERC Starting Grant 758800 (EXPROTEA), ERC Consolidator Grant 101087347 (VEGA), ANR AI Chair AIGRETTE, as well as gifts from Ansys and Adobe Research. This work was also supported by the Galileo 2022 fellowship from the Università Italo Francese/ Université Franco Italienne (UIF/UFI) within the project  G22\_4 titled “Multimodal Artificial Intelligence for 3D shape analysis, modeling and applications“.

{\small
\bibliographystyle{ieeenat_fullname}
\bibliography{11_references}
}

\ifarxiv \clearpage \appendix \maketitlesupplementary
\setcounter{figure}{7}
\setcounter{table}{1} 

\begin{table*}[t]
\centering
\resizebox{\linewidth}{!}{%
\begin{tabular}{c|c|cc|cc|cc}
\hline
\textbf{Method} & \textbf{3D Encoder} & \multicolumn{2}{c|}{\textbf{CLIP}} & \multicolumn{2}{c|}{\textbf{RoBERTa}} & \multicolumn{2}{c}{\textbf{BERT}} \\ 
                &                     & Top-1 retrieval & Top-10 retrieval & Top-1 retrieval & Top-10 retrieval & Top-1 retrieval & Top-10 retrieval \\ \hline
\multicolumn{8}{c}{\textbf{Multi-modal 3D encoder}} \\ \hline
Affine + Subspace Projection     & OpenShape           & 56.4            & 90.8            & 38.8            & 81.8            & 32.4            & 78.8            \\ 
Affine + Subspace Projection     & ULIP-2           & 54.2            & 90.2            & 37.0            & 80.8            & 29.2            & 69.2            \\ 
Affine + Subspace Projection     & Uni3D              & 47.0            & 89.0            & 29.4            & 73.8            & 19.0            & 59.4            \\ \hline
\multicolumn{8}{c}{\textbf{Uni-modal 3D encoder}} \\ \hline
Affine       & PointBert           & 9.8            & 37.2            & 42            & 22.2            & 3.4             & 22.6            \\ 
Affine       & SparseConv              & 10.6            & 46.2            & 8.0             & 29.4            & 3.2             & 20.4            \\ 
Affine       & Pointnet            & 7.0            & 30.2            & 3.4             & 22.0            & 3.0             & 20.0            \\ \hline
Affine + Subspace Projection (Ours)       & PointBert           & 18.0            & 57.4            & 10.8            & 36.6            & 7.6             & 25.8            \\ 
Affine + Subspace Projection (Ours)       & SparseConv              & 13.0            & 58.0            & 8.2             & 29.4            & 5.2             & 21.0            \\ 
Affine + Subspace Projection (Ours)       & Pointnet            & 14.0            & 44.8            & 7.0             & 35.8            & 6.6             & 23.8  \\ \hline
Local CKA       & PointBert           & 5.4            & 24.4            & 0.2            & 4.0            & 0.8             & 7.19            \\ 
Local CKA       & SparseConv              & 3.4            & 23.59            & 0.2             & 3.2          & 0.6            & 6.4            \\ 
Local CKA       & Pointnet            & 4.2            & 28.19            & 0.0             & 3.59            & 1.0            & 8.0  \\ \hline
Local CKA + Subspace Projection (Ours)       & PointBert           & \textbf{30.0}            & \textbf{70.8}            & \textbf{17.8}            & \textbf{54.4}            & 13.6             & \textbf{51.0}            \\ 
Local CKA + Subspace Projection (Ours)       & SparseConv              & 21.2            & 64.0            & 14.79             & 42.4            & 11.4             & 41.4           \\ 
Local CKA + Subspace Projection (Ours)       & Pointnet            & 23.79            & 62.6            & 15.8             & 49.2            & \textbf{14.6}             & 45.4 \\ \hline
\end{tabular}%
}
\caption{\textbf{Top-1 and top-5 retrieval accuracy across 3D and text encoders using different alignment approaches.} We use 30,000 anchors for subspace projection and affine transformation approaches, and 1,000 anchors for local CKA. A query set of 500 is uniformly sampled, with results averaged over 3 different seeds. The subspace dimension is fixed at 50. Our approach (Ours) consistently demonstrates improved retrieval performance, with multi-modal 3D encoders setting the upper bound for performance.}
\vspace{-0.05in}
\label{tab:performance_supp}
\end{table*}

\begin{table*}[t]
\centering
\resizebox{\linewidth}{!}{%
\begin{tabular}{c|c|cc}
\hline
\textbf{Method} & \textbf{3D Encoder} & \multicolumn{2}{c}{\textbf{T5}} \\ 
                &                     & Matching accuracy & Top-5 retrieval \\ \hline
\multicolumn{4}{c}{\textbf{Multi-modal 3D encoder}} \\ \hline
Affine + Subspace Projection (Ours)     & OpenShape           & 65.0            & 82.6             \\ 
Affine + Subspace Projection (Ours)     & ULIP-2              & 51.8            & 73.2             \\ 
Affine + Subspace Projection (Ours)     & Uni3D               & 53.6            & 67.0             \\ \hline
\multicolumn{4}{c}{\textbf{Uni-modal 3D encoder}} \\ \hline
Affine + Subspace Projection (Ours)     & PointBert           & 21.6            & 28.4             \\ 
Affine + Subspace Projection (Ours)     & SparseConv          & 22.8             & 23.2             \\ 
Affine + Subspace Projection (Ours)     & Pointnet            & 21.6             & 22.0             \\ \hline
\end{tabular}%
}
\caption{\textbf{Matching and Top-5 retrieval accuracy using T5 text encoder and different 3D encoders.} The Affine + Subspace Projection (Ours) method is evaluated across both multi-modal and uni-modal 3D encoders. T5 is better aligned to 3D encoders compared to other uni-modal text encoders as presented in Table 1 of the main paper.}
\vspace{-0.05in}
\label{tab:t5_performance}
\end{table*}

This supplementary document provides additional technical details and experimental results to complement the main material. Specifically, we first outline some technical details pertinent to our framework in \cref{sec:technical_details}. Then, we evaluate in \cref{sec:text_encoder} the alignment performance using a different text encoder, T5. Next, we extend the matching and top-5 downstream results from Table 1 in the main paper by including top-1 and top-10 retrieval scores in \cref{sec:top1and10}. Furthermore, in \cref{sec:ablation}, we examine the impact of chosen subspace dimensions across different pairs of text and 3D encoders. These additional results provide a more comprehensive understanding of our alignment approach and its robustness across different configurations.

\section{Implementation Details}
\label{sec:technical_details}

In our framework, we use mean pooling for text encoders to obtain a fixed-size text representation. While we also experimented with using a class token for text encoding, it yielded consistently similar results. For 3D encoders, we extract the global output feature as the final representation.

The embedding size is set to 512 whenever possible, with the option to use projection layers when necessary. However, for multi-modal pre-trained models and certain text encoders, the embedding dimension may vary. In such cases, we adopt two distinct strategies: (1) for Canonical Correlation Analysis (CCA)-based approaches (Ours), the maximum subspace dimension is determined as the minimum of the embedding sizes of the 3D and text encoders; (2) for affine transformation-based alignment, we follow prior work by padding the lower-dimensional representation to match the higher-dimensional one.

For training parameters and dataset configurations related to uni-modal encoders, we adhere to the OpenShape settings, ensuring consistency with existing benchmarks.

\section{Evaluation with an Additional Text Encoder}
\label{sec:text_encoder}

To further evaluate the generalization of our alignment approach, we test it using an additional text encoder, T5. Unlike BERT, RoBERTa, and CLIP’s text encoder, which are encoder-only architectures, T5 follows an encoder-decoder structure. We average its encoder output embeddings and use the resulting vector as the text representation.

As shown in \cref{tab:t5_performance}, while T5 does not achieve the same performance as CLIP’s text encoder, it consistently outperforms other uni-modal text encoders, such as BERT and RoBERTa, in alignment tasks. These results suggest that the encoder-decoder structure may provide richer text representations for cross-modal alignment, although multi-modal encoders like CLIP text encoder remain superior for this task.

\section{Downstream Results}
\label{sec:top1and10}

We extend our evaluation by including top-1 and top-10 retrieval metrics, which complement the matching and top-5 results presented in the main paper by offering additional perspectives. As shown in \cref{tab:performance_supp}, these results emphasize the consistency of our findings: the combination of local CKA and our proposed subspace projection method consistently achieves superior performance in retrieval tasks, whereas the affine approach demonstrates better results in matching tasks (Table 1). This highlights that method performance can vary significantly depending on the downstream task, reflecting the distinction between overall assignment accuracy (matching) and query-specific precision (top-1 retrieval). Among uni-modal 3D encoders, PointBERT performs best. Meanwhile, CLIP continues to excel as the most effective text encoder, which shows its generalizability across modalities.

The alignment approaches studied thus far exhibit limited generalization to the zero-shot classification downstream task. In particular, top-1 accuracy on Objaverse-LVIS remains below 3\% even for the best-performing uni-modal 3D encoders when aligned with text encoders which is way lower to the 40\% and more attained by OpenShape. This performance bottleneck can be attributed primarily to the repetitive nature of the captions used in this task: instances within the same class often share identical or nearly identical textual descriptions, leading to duplicated text embeddings. This opens up a new direction to enhance these approaches with zero-shot classification capabilities.

\begin{figure*}[tp]
    \centering
    \begin{subfigure}[b]{0.32\textwidth}
        \centering
        \includegraphics[width=\textwidth]{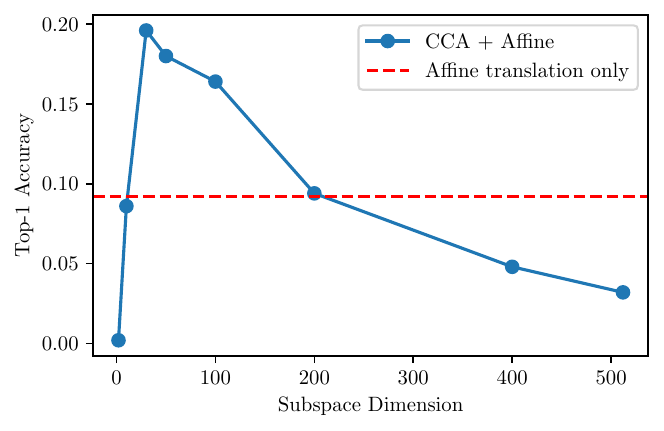}
        \caption{CLIP and PointBert}
        \label{fig:pair1}
    \end{subfigure}
    \hfill
    \begin{subfigure}[b]{0.32\textwidth}
        \centering
        \includegraphics[width=\textwidth]{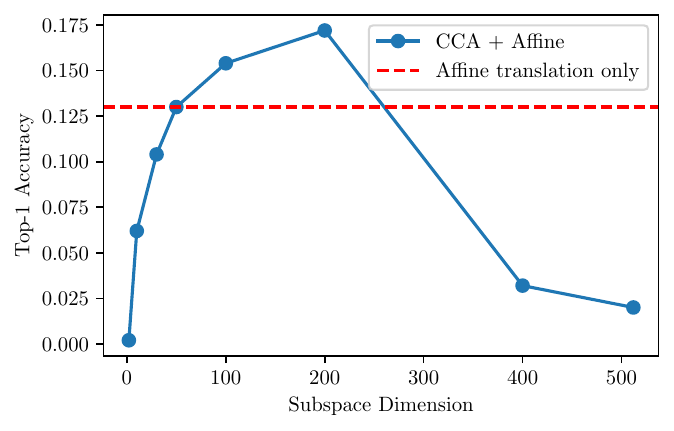}
        \caption{CLIP and MinkowskiNet}
        \label{fig:pair2}
    \end{subfigure}
    \hfill
    \begin{subfigure}[b]{0.32\textwidth}
        \centering
        \includegraphics[width=\textwidth]{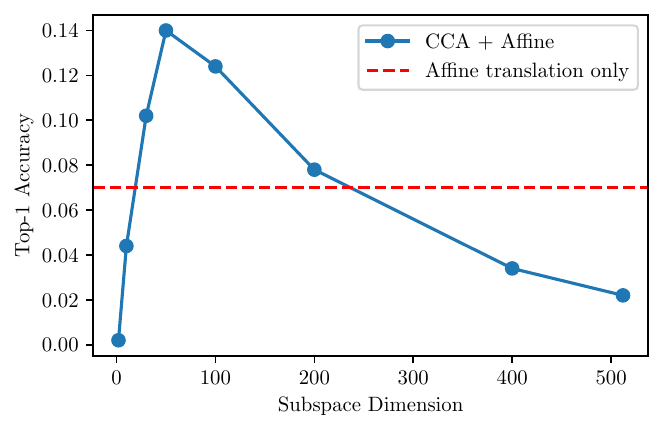}
        \caption{CLIP and PointNet}
        \label{fig:pair3}
    \end{subfigure}

    \vspace{0.5cm}
    
    \begin{subfigure}[b]{0.32\textwidth}
        \centering
        \includegraphics[width=\textwidth]{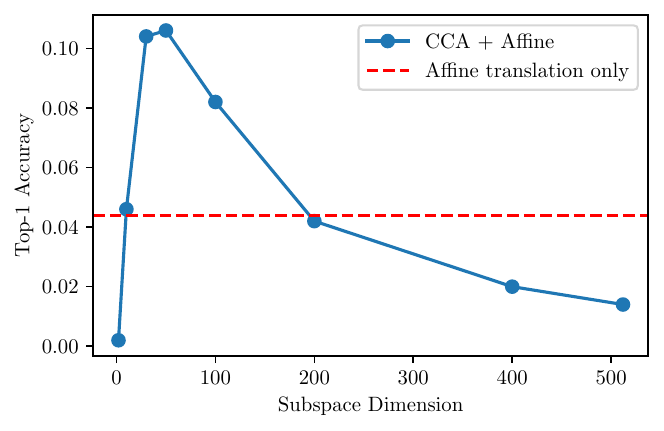}
        \caption{RoBERTa and PointBert}
        \label{fig:pair4}
    \end{subfigure}
    \hfill
    \begin{subfigure}[b]{0.32\textwidth}
        \centering
        \includegraphics[width=\textwidth]{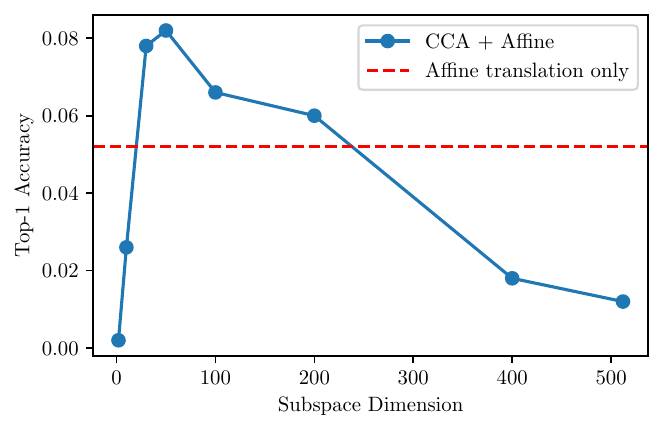}
        \caption{RoBERTa and MinkowskiNet}
        \label{fig:pair5}
    \end{subfigure}
    \hfill
    \begin{subfigure}[b]{0.32\textwidth}
        \centering
        \includegraphics[width=\textwidth]{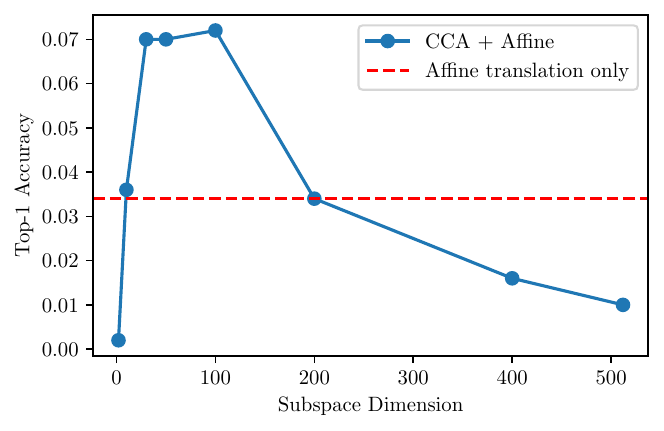}
        \caption{RoBERTa and PointNet}
        \label{fig:pair6}
    \end{subfigure}

    \vspace{0.5cm}
    
    \begin{subfigure}[b]{0.32\textwidth}
        \centering
        \includegraphics[width=\textwidth]{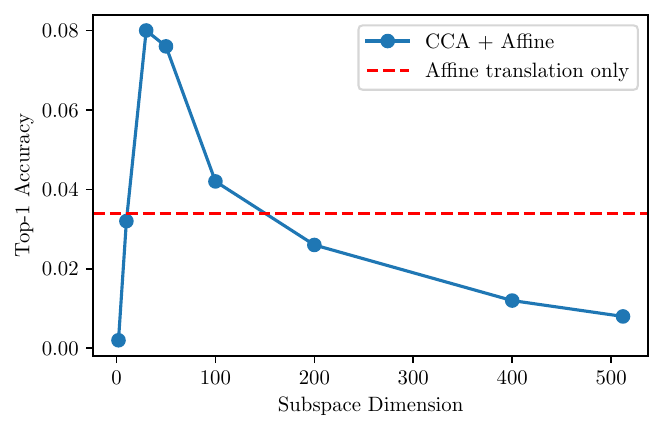}
        \caption{BERT and PointBert}
        \label{fig:pair7}
    \end{subfigure}
    \hfill
    \begin{subfigure}[b]{0.32\textwidth}
        \centering
        \includegraphics[width=\textwidth]{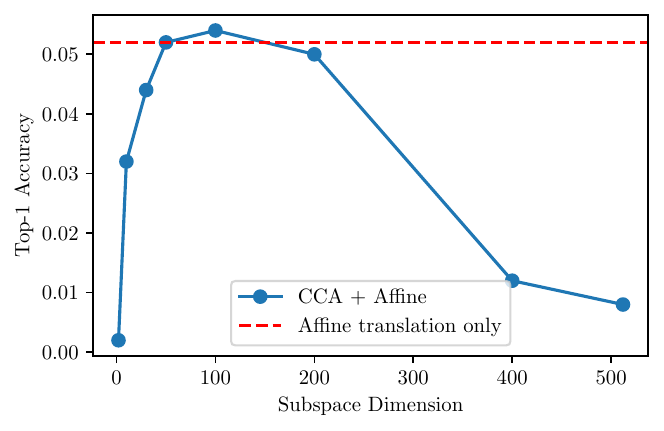}
        \caption{BERT and MinkowskiNet}
        \label{fig:pair8}
    \end{subfigure}
    \hfill
    \begin{subfigure}[b]{0.32\textwidth}
        \centering
        \includegraphics[width=\textwidth]{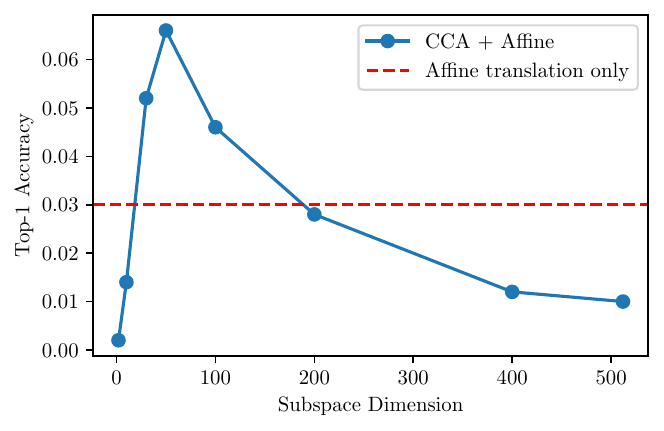}
        \caption{BERT and PointNet}
        \label{fig:pair9}
    \end{subfigure}

    \caption{\textbf{Impact of subspace dimensionality on retrieval performance.} Comparison of two approaches: our proposed CCA + affine translation method (blue) and affine translation without subspace projection (red). Each plot corresponds to a pair of Text Encoder and 3D Encoder. Optimal downstream performance is obtained with low-dimensional subspace projection, although the exact dimension differs from encoder to another.}
    \label{fig:dimension_ablation_all}
\end{figure*}

\section{Additional Ablations}
\label{sec:ablation}

\mypara{Dimensionality's impact on alignment.} We generalize the dimension analysis to additional pairs of text and 3D encoders in \cref{fig:dimension_ablation_all}, extending the findings presented in the main paper. The results confirm that our method consistently achieves better alignment in low-dimensional subspaces across all evaluated pairs, which reaffirms the importance of dimensionality reduction to enable our subspace projection approach. The optimal subspace dimension is often consistent across different encoders, but exceptions are observed. For example, MinkowskiNet exhibits improved performance at higher dimensions (e.g. 200 vs. 50), which shows that encoders have representations that might align differently. This variability highlights that the ideal subspace dimension for balancing geometric and semantic features, while being low, is not fixed but encoder-dependent.

\mypara{Experiments Favoring Uni-Modal Encoders:} Although our focus is on bridging the gap between uni-modal latent spaces, we can demonstrate that uni-modal 3D encoders have an edge over multi-modal encoders in geometry understanding. For instance, the Pearson correlation experiment (\cref{fig:pearson_multimodal} which complements Fig. 6 in the main) shows that the subspaces emerging from text 3D multi-modal encoders do not correlate with geometric similarity, \textit{thus potentially limiting their utility} in tasks, where geometric similarity is important.

\begin{figure}[tp]
    \centering
    \includegraphics[width=\linewidth]{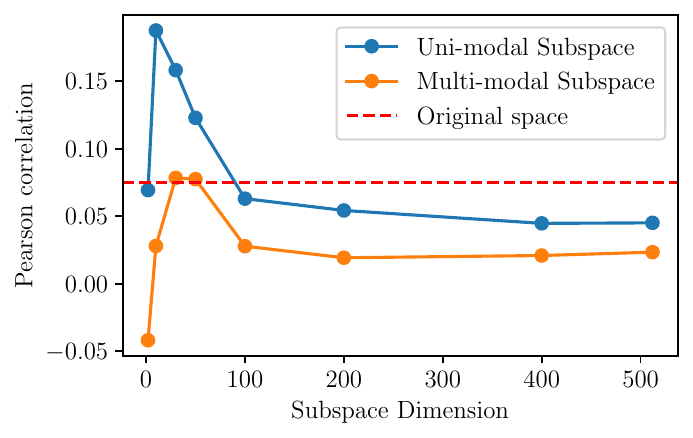}
    \caption{\textbf{Pearson correlation between shape query Chamfer distances and pairwise distances.} We observe a higher Pearson correlation in the projected text uni-modal latent subspace compared to the projected text multi-modal subspace.}
    \label{fig:pearson_multimodal}
\end{figure} \fi

\end{document}